\documentclass[journal]{IEEEtran}
\usepackage[T1]{fontenc}
\usepackage{aecompl}
\usepackage{amsmath,amssymb,amsfonts}
\usepackage{mathrsfs}
\usepackage{array}
\usepackage{subfigure}
\usepackage{textcomp}
\usepackage{stfloats}
\usepackage{url}
\usepackage{verbatim}
\usepackage{graphicx}
\usepackage{multirow}
\usepackage{amsthm}
\usepackage[linesnumbered,ruled,vlined]{algorithm2e}
\usepackage{bbm}
\usepackage{hyperref}
\usepackage{orcidlink}
\usepackage{cite}
\usepackage{booktabs}
\newcommand{\ie}{\textit{i}.\textit{e}.,}

\begin{document}

\title{\bf \LARGE Multi-Robot Reliable Navigation in Uncertain Topological Environments with Graph Attention Networks}

\author{Zhuoyuan Yu, Hongliang Guo, Albertus Adiwahono, Jianle Chan, Brina Shong, Chee-Meng Chew, Wei-Yun Yau %
        % <-this % stops a space
\thanks{Manuscript received in November 2024; This work was supported in part by the Robotics Horizontal Technology Coordinating Office, Agency for Science, Technology and Research, Singapore, under Grant C221518004;}% <-this % stops a space
\thanks{Zhuoyuan Yu, Hongliang Guo, Albertus Adiwahono, Jianle Chan, Brina Shong, and Wei-Yun Yau are with the Institute for Infocomm Research (I2R), Agency for Science, Technology and Research (A*STAR), Singapore 138632.}
% (e-mail: \texttt{yuzhuoyuan@u.nus.edu}; \texttt{\{guo\_hongliang, wyyau, chan\_jian\_le, brina\_shong\}@i2r.a-star.edu.sg.})}
\thanks{Zhuoyuan Yu and Chee-Meng Chew are with the College of Design and Engineering of National University of Singapore (NUS), Singapore 138600.}}

% The paper headers
% \markboth{IEEE ROBOTICS AND AUTOMATION LETTERS, VOL.X, NO.X, XXXX 2024}
% {Shell \MakeLowercase{\textit{et al.}}: A Sample Article Using IEEEtran.cls for IEEE Journals}

% \IEEEpubid{2377-3766\copyright~2024 IEEE.Personal use is permitted, but republication/redistribution requires IEEE permission.\\See https://www.ieee.org/publications/rights/index.html for more information.}
% Remember, if you use this you must call \IEEEpubidadjcol in the second
% column for its text to clear the IEEEpubid mark.

\maketitle
\pagestyle{empty}  % no page number for the second and the later pages
\thispagestyle{empty} % no page number for the first page
\begin{abstract}
% This research addresses the challenge of reliable multi-robot navigation in uncertain topological networks, to maximizing the probability of on-time arrival despite unpredictable conditions. The uncertainty in these networks stems from unknown edge traversability, which is only revealed when robots arrive at the corresponding nodes. Existing approaches often struggle to adapt to real-time changes, making them unsuitable for dynamic environments. To overcome this issue, we reformulate the problem within a Decentralized Partially Observable Markov Decision Process (Dec-POMDP) framework and introduce the Dynamic Adaptive Graph Embedding method to better capture the evolving nature of the navigation task. We further enhance policy learning by integrating deep reinforcement learning with Graph Attention Networks (GATs), leveraging self-attention to focus on critical graph features. Our approach, Multi-Agent Routing in Variable Environments with Learning (MARVEL) employs an online reinforcement learning algorithm to iteratively optimize real-time decision-making. We validate the effectiveness of MARVEL through comprehensive simulations, demonstrating improved adaptability and performance in uncertain networks. Additionally, real-world experiments with robotic teams navigating within indoor and topological transportation networks confirm the practical applicability of MARVEL.
This paper studies the multi-robot reliable navigation problem in uncertain topological networks, which aims at maximizing the robot team's on-time arrival probabilities in the face of road network uncertainties. The uncertainty in these networks stems from the unknown edge traversability, which is only revealed to the robot upon its arrival at the edge's starting node. Existing approaches often struggle to adapt to real-time network topology changes, making them unsuitable for varying topological environments. To address the challenge, we reformulate the problem into a Partially Observable Markov Decision Process (POMDP) framework and introduce the Dynamic Adaptive Graph Embedding method to capture the evolving nature of the navigation task. We further enhance each robot's policy learning process by integrating deep reinforcement learning with Graph Attention Networks (GATs), leveraging self-attention to focus on critical graph features. The proposed approach, namely Multi-Agent Routing in Variable Environments with Learning (MARVEL) employs the generalized policy gradient algorithm to optimize the robots' real-time decision-making process iteratively. We compare the performance of MARVEL with state-of-the-art reliable navigation algorithms as well as Canadian traveller problem solutions in a range of canonical transportation networks, demonstrating improved adaptability and performance in uncertain topological networks. Additionally, real-world experiments with two robots navigating within a self-constructed indoor environment with uncertain topological structures demonstrate MARVEL's practicality.
\end{abstract}

\section{Introduction}
\label{Intro}
\IEEEPARstart{A}{dvancements} in autonomous systems have made multi-robot navigation a vital research area, particularly when navigating in dynamic environments. In real-world scenarios, robots often encounter situations which require decisions to be made under uncertain conditions. Such uncertainty may stem from various factors, including randomness in indoor environments or fluctuating accessibility due to weather conditions in outdoor environments \cite{guo2019robust}. This challenge is especially relevant to intelligent transportation systems, drawing growing attention from academia and industry.

To address this challenge, various algorithms have been developed. Existing solutions are typically categorized by different objective functions. A widely used approach is navigation based on the least expected time, which is suitable for scenarios where occasional delays are acceptable, such as daily commuting. Under different conditions, ensuring on-time arrival probability becomes a critical priority, such as in emergency response or time-sensitive deliveries \cite{cao2020using}.

\begin{figure}[!t]
\centering
\includegraphics[width=3in]{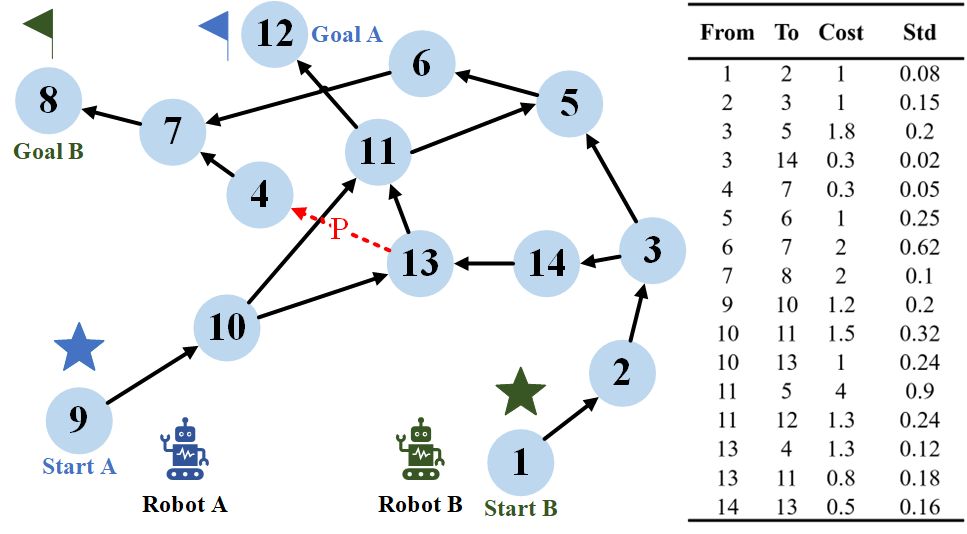}
\caption{The illustrative example: In the topological network, the blue and green markers represent Robot A and Robot B, respectively. The red dashed edge is an uncertain edge labelled with traversal probability. The right table provides the corresponding data for the topological map. Each edge has an average cost and standard deviation; the actual cost for each edge is sampled from a Gaussian distribution with rejection. The \textbf{objective} is to maximize the on-time arrival probability for the entire team, which is computed based on the data from subsequent edges combined with time budget $T_i$. In other algorithms, Robot A always selects 9→10→11→12, and Robot B typically learns a fixed policy to select either 5 or 14 at node 3, regardless of real-time conditions. However, MARVEL guides Robot A to take path 9→10→13→11→12 to test the traversability of edge 13→4, allowing adaptive decision-making for Robot B at node 3 based on the real-time transportation networks.}
\label{fig_1}
\vspace{-5pt}
\end{figure}

However, current navigation approaches on topological maps primarily focus on single-agent settings, leading to several limitations. Firstly, these methods rely on prior knowledge of edge traversability probabilities, which is rarely achievable in real-world environments where disruptions are often unpredictable. Additionally, the paths generated by these algorithms are predetermined before departure, reducing their adaptability to dynamic conditions. As illustrated in Fig.~\ref{fig_1}, if the traversal probability of edge 13-4 is high, Robot B will be directed to node 14. However, if this segment becomes impassable during actual movement, the cost of reaching node 8 will significantly increase, making delays almost inevitable. Conversely, if the probability is low, or for risk-averse algorithms, Robot B will be consistently directed to node 5, inevitably increasing the average travel cost. These limitations reduce the applicability of such algorithms in complex, real-world environments.

\IEEEpubidadjcol
In response to these challenges, we propose Multi-Agent Routing in Variable Environments with Learning (MARVEL), a framework which is designed to operate without relying on prior knowledge of edge passability. Unlike other methods, MARVEL enables real-time planning and rerouting, allowing robots to plan routes based on the explored map, thus significantly mitigating potential risks due to environmental changes. We refine existing graph embedding methods to better handle uncertainty and utilize an online expert as a real-time planner to guide the process. This approach enables lower-priority robots to sacrifice some probability of on-time arrival in favor of meaningful exploration, thereby enhancing the overall on-time arrival probability for the team.

The main contributions of this research are as follows: (1) Proposing a novel multi-agent algorithm for navigating uncertain topological networks: We introduce a new method that leverages team collaboration to improve the reliability and robustness of navigation policies across multi-robot systems; (2) Developing a dynamic adaptive graph embedding algorithm for variable networks: MARVEL updates node features based on the real-time road networks, significantly enhancing adaptability and robustness in dynamic topological networks; (3) Integrating Graph Attention Networks with the entropy-based online expert: MARVEL leverages this combination to focus on critical edges, improving decision-making robustness and enhancing adaptability in complex environments.

This paper is organized as follows. In Section \ref{Literature Review}, we provide a succinct review of the relevant literature. Section \ref{problem_formulation} covers the problem formulation and the related background. Our proposed methodology is then detailed in Section \ref{MA-CTP}. Section \ref{Simulation Results and Analysis} presents the evaluation and comparison of MARVEL against the existing advanced approaches. Section \ref{Deployment and Experiment Results} describes the real-world deployment of MARVEL onto physical robots for experimental validation. Finally, Section \ref{Conclusion and Future Work} concludes this research with a discussion of potential future work.

\section{Literature Review}
\label{Literature Review}
The Multi-Agent Canadian Traveller Problem (MA-CTP) is a multi-agent navigation problem where agents need to obtain an advantageous policy in a topological network with unknown information. As each agent is navigated through the topological network, it is crucial to balance selfish and altruistic behaviors effectively. Based on our research, although there are studies that attempt to address the Multi-Agent Path Finding (MAPF) problem with specific deadlines \cite{wang2022multi, yang2024attention} and on-time arrival probabilities \cite{cao2016multiagent, ma2017multi}, there is lack of research combining CTP with multi-agent systems. The few existing works that integrate CTP with multi-agent algorithms \cite{shiri2017online} primarily focus on agents sharing the same start and end points. These studies differ from conventional multi-agent algorithms and rely heavily on methods similar to single-agent algorithms, lacking team collaboration and failing to reduce risk. This distinguishes our research significantly.

\subsection{Canadian Traveller Problem}

The Canadian Traveller Problem (CTP), initially proposed in 1991 \cite{papadimitriou1991shortest}, involves determining the optimal policy under stochastic networks which have some edges with unknown passable probability, with this information being revealed only upon the robot's arrival at an incident vertex of uncertain edges\cite{guzzi2019impact}. Current research on CTP primarily distinguishes between different objective functions outlined below:

\subsubsection{Least Expected Time (LET)}This approach aims to find the path with the minimum average cost \cite{liao2015generalized}. It is the most common and widely applicable method. As mentioned in the previous section, it is suitable for such scenarios where saving the average expected time is of great importance, and some delay is acceptable, such as daily commuting.

\subsubsection{Minimal Worst-Case Occurrence} This approach prioritizes the avoidance of scenarios where the destination becomes unreachable or the travel time significantly exceeds the average \cite{nikolova2008route}. Rather than focusing on strict arrival times, this method aims to ensure that worst-case outcomes are avoided, making it well-suited for tasks involving non-urgent transport.

\subsubsection{Minimal Mean-Var Combination} This approach integrates travel time with the variance of each segment to maximize the likelihood of on-time arrival \cite{guo2019robust}. In recent years, this approach has emerged as a major focus due to its effectiveness in optimizing the probability of timely arrival. However, as mentioned before, these methods lack real-time adaptability and are unable to respond to changes in real-time conditions.

Furthermore, CTP also has several variants that can be classified based on different criteria. For instance, the k-CTP restricts the total number of blocked edges \cite{bender2015optimal}, while the recoverable CTP allows blocked edges to become passable after a certain time \cite{su2009recoverable}. Another variant is the Multi-Target CTP, which involves visiting a set of targets\cite{liao2014covering}. These examples are just a few of the many variants that exist.

In contrast to the previous solutions, MARVEL tackles the CTP through multi-agent collaboration, enabling agents to make dynamic decisions based on continuous interaction. By incorporating an entropy-enhanced objective function, MARVEL further facilitates real-time information exchange, optimizing decision-making in uncertain environments.

\subsection{Multi-Agent Path Finding}
The increasing deployment of multi-agent systems in domains such as autonomous driving \cite{okoso2021network}, multi-robot coordination \cite{ma2021distributed, he2024social}, and even gaming \cite{ma2017feasibility} lead to the development of numerous advanced algorithms to tackle the Multi-Agent Path Finding (MAPF) problem. A significant portion of these studies concentrates on devising cost-efficient and conflict-free solutions within grid maps, a critical requirement for applications such as transportation, and warehousing\cite{alkazzi2024comprehensive}.

State-of-the-art research on MAPF algorithms primarily focuses on learning-based approaches\cite{he2024alpha}. Many of these methods have been applied to 1) distributed reinforcement learning techniques like $\text{Primal}_2$\cite{damani2021primal}, as well as 2) multi-agent reinforcement learning (MARL) frameworks such as SCRIMP\cite{wang2023scrimp} and PICO\cite{li2022multi}, which often integrate MARL with traditional neural networks. Recently, graph neural networks (GNNs) have gained prominence for their ability to process graph-structured and irregular data, making them increasingly applicable to MAPF tasks \cite{li2020graph}. Among these, Graph Attention Networks (GATs), which leverage a self-attention mechanism to dynamically weigh the importance of vertices, have proven effective not only in static graphs but also in more complex environments, such as large-scale road networks, aligning seamlessly with the demands of MAPF scenarios \cite{li2021message}.

Despite significant advancements in MAPF research, much of the focus remains on grid maps\cite{nguyen2020deep}, with limited attention given to topological networks. This is largely due to the scalability challenges\cite{kiran2021deep}, as MAPF on grid maps is already an $NP$-hard problem, and the complexity becomes more pronounced while extended to topological networks. Nevertheless, this does not deter us from exploring the potential of applying GAT in dynamic transportation networks. In this paper, we experimentally validate the use of GAT in dynamic graphs and further optimize existing multi-agent navigation methods.

\section{Problem Formulation and Backgrounds}
\label{problem_formulation}

This section first outlines the problem formulation, establishing the foundation for MARVEL. We then provide concise overviews of the relevant techniques in MARVEL, including the Skip-Gram algorithm for generating embeddings, information entropy to quantify uncertainty, and Graph Convolutional Networks for capturing the underlying structure of graphs.

\subsection{Problem Formulation}
The problem can be reformulated as a Partially Observable Markov Decision Process (POMDP). To formalize and comprehensively describe this problem, we use two tuples $\langle \mathcal{G} = (\mathcal{V}, \mathcal{E}), \mathcal{C}\rangle$, $\langle \mathcal{R}, (\mathcal{O}, \mathcal{D}), \mathcal{T}, \mathcal{I} \rangle$ to represent the graph structure and agent characteristics respectively. Only the main symbols are listed here. For a complete list of symbols, please refer to the table provided in the code documentation.

\begin{itemize}
    \item $\mathcal{G} = (\mathcal{V}, \mathcal{E})$ represents a directed graph, with $\mathcal{V}$ being the set of all nodes and $\mathcal{E}$ being the set of all edges (including traversable and probabilistic edges).
    \item $\mathcal{C}$: $e \in \mathcal{E} \rightarrow \mathbb{R^+} $ calculates the impact on the whole team's probability of arriving on time at the destination after a robot chooses a particular edge. This includes information entropy and the impact on the team's arrival probability.
    \item $\mathcal{R}, (\mathcal{O}, \mathcal{D}), \mathcal{T}, \mathcal{I}$ are the sets representing the robots, with their corresponding origin nodes, destination nodes, time budget and the importance of tasks, respectively. 
\end{itemize}

The objective of this problem is to find the policy:
\begin{equation}\label{eq_1}
\boldsymbol{\phi}^* = \arg\max_{\boldsymbol{\phi} \in \Phi}\left\{ \sum\limits_{i \in \mathcal{A}}\lambda_i \times \mathcal{P}_i(\boldsymbol{\phi}) - \mathbb{H}(\mathcal{G}) \right\},
\end{equation}

where $\Phi$ is the set denoting all the possible policies, $\lambda_i$ represents the weight of importance, subject to the conditions $\lambda_i \in (0, 1) \quad \text{and} \quad \sum_{i=1}^n \lambda_i = 1$,
and $\mathcal{P}_i(\boldsymbol{\phi})$ denotes the impact on each agent's probability of arriving on time.

\subsection{Graph Embedding: Skip-Gram}

To develop a scalable solution, we employ the graph embedding technique to convert the roadnet into $d$-dimensional vectors. MARVEL utilizes the skip-gram method, commonly used in Natural Language Processing (NLP), to compress nodes into feature vectors efficiently. It is an unsupervised learning technique that captures semantic information by learning relationships between words and their contexts \cite{mikolov2013efficient}.

Given a sequence of words \( w_1, w_2, \ldots, w_T \), the word is indexed as \(i\) in the dictionary. When it serves as a center word, it is represented by the vector \(\boldsymbol{v_i} \in \mathbb{R}^d\), while as a context word, it is represented by the vector \(\boldsymbol{u_i} \in \mathbb{R}^d\). Let the center word \(w_c\) be indexed as \(c\), and the context word \(w_o\) be indexed as \(o\). The conditional probability of generating the context word given the center word can be obtained by applying a SoftMax operation to the inner product of the vectors:

\begin{equation}\label{eqn-2}
\mathbb{P}(w_{o} \! \mid \! w_{c}) = \frac{\exp(\boldsymbol{u^\top_0} \cdot \boldsymbol{{v}_c})}{\sum_{i\in\mathcal{W}} \exp(\boldsymbol{u^\top_i} \cdot \boldsymbol{v_c})}.
\end{equation}
Given that the words are assumed to be independent, the overall probability is the product of individual probabilities:

\begin{equation}\label{eqn-1}
    \prod_{t=1}^{T} \prod_{\substack{-m \leq j \leq m , j \neq 0}} P(w^{(t+j)} \mid w^{(t)}),
\end{equation}
where \( T \) is the position of the window's center word, and \( m \) denotes the size of the window. This allows us to compute the probability of inferring the context words for center words. 

\subsection{Information Entropy}

Information entropy measures the uncertainty associated with a variable. It represents the average amount of information produced by stochastic data\cite{shannon1948mathematical}. To quantify the uncertainty of the road networks, information entropy is incorporated into the objective function, guiding the agents to explore.

For the discrete random variable $ \boldsymbol{X} $ with possible values $ x_1, x_2, \ldots, x_n $ and corresponding probabilities $\mathbb{P}(x_1)$, $\mathbb{P}(x_2)$, $\ldots, \mathbb{P}(x_n)$ , the information entropy $ \mathbb{H}(\boldsymbol{X}) $ is defined as:

\begin{equation}\label{eqn-3}
\mathbb{H}(\boldsymbol{X}) = -\sum_{i=1}^{n} \mathbb{P}(x_i) \log \mathbb{P}(x_i).
\end{equation}
From Eq.~\eqref{eqn-3}, we can see that higher entropy corresponds to greater uncertainty and higher information content, which will be employed in subsequent sections to guide the policy.

\subsection{Graph Convolutional Networks}

Graph Convolutional Network (GCN) is a type of neural network designed specifically for processing graph-structured data. By aggregating information from the neighbourhood of each node, GCNs capture local structural information like how convolutional operations work in images\cite{kipf2016semi}. 

GCN could be defined on the given graph \( \mathcal{G} = (\mathcal{V}, \mathcal{E}) \). From this, we can derive the \textbf{node feature matrix} \( \boldsymbol{X} \in \mathbb{R}^{{\mathcal{N}} \times \mathcal{F}} \), where \( \mathcal{N} \) is the number of nodes and \( \mathcal{F} \) is the number of features per node; and the \textbf{adjacency matrix} \( \boldsymbol{A} \in \mathbb{R}^{{\mathcal{N}} \times \mathcal{N}} \), which is used to represent the connections between nodes. Then we normalize the adjacency matrix as follows:

\begin{equation}\label{eqn-4}
{\boldsymbol{\hat{A}}} = \boldsymbol{D}^{-\frac{1}{2}} (\boldsymbol{A} + \boldsymbol{I}) \boldsymbol{D}^{-\frac{1}{2}},
\end{equation}
where $\boldsymbol{D}$ is the degree matrix, and diagonal elements correspond to the degree of each node. This step stabilizes training. Once normalized, the convolution operation is performed as:

\begin{equation}\label{eqn-5}
\boldsymbol{F}^{(l+1)} = \sigma(\boldsymbol{\hat{A}}\boldsymbol{F}^{(l)} \boldsymbol{W}^{(l)}),
\end{equation}
where $\boldsymbol{F}^{(l)}$ represents the features at layer $l$, $\boldsymbol{W}^{(l)}$ is a trainable weight matrix at layer \( l \), and \( \sigma \) is the activation function. After multiple layers of convolution, GCN iteratively aggregates and combines information from neighbours.

\section{Multi-Agent Routing in Variable Environments with Learning}
\label{MA-CTP}

\begin{figure*}[t] 
\centering
\includegraphics[width=\textwidth]{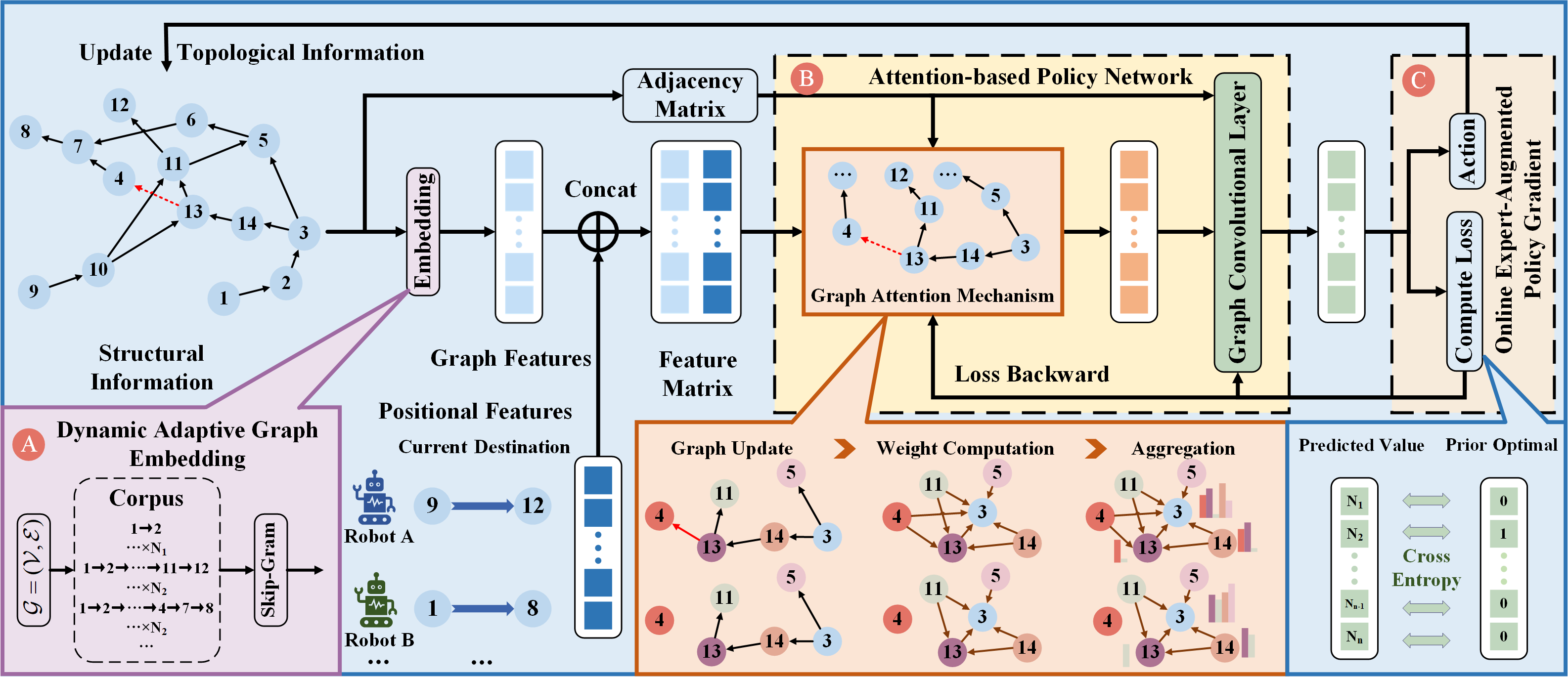} 
\caption{\textbf{The MARVEL framework} outlines a processing pipeline composed of a graph embedder, a graph neural network module, and a policy update module based on the online expert. (i) The feature matrix and adjacency matrix are dynamically updated by the real-time structural and positional information. The corpus is generated for each node based on neighbors and shortest paths to all destination nodes. (ii) Our policy network captures topological features, calculates weights using a self-attention mechanism, and selectively aggregates useful information for decision-making. The bar chart represents the influence weights of subsequent vertices in different scenarios. (iii) The policy is dynamically optimized by the online expert-augmented policy gradient method. The cross-entropy loss is calculated between predicted values and prior solutions, enabling effective adjustment of the attention distribution and policy updates.}

\label{pipeline}
\end{figure*}

This section outlines the MARVEL methodology for the multi-robot navigation problem in uncertain topological road networks, as illustrated in Fig.~\ref{pipeline}. We first propose an adaptive graph embedding operation tailored for navigation tasks in dynamic transportation networks. This is followed by a policy network based on Graph Attention Networks (GATs), which leverages the self-attention mechanism to dynamically assign weights to neighboring nodes based on real-time road conditions, guiding decision-making accordingly. The training process is conducted through online expert supervision, where prior optimal solutions are utilized to guide the policy network, functioning as a heuristic to reduce ineffective search in large-scale networks and improve convergence efficiency.

\subsection{Dynamic Adaptive Graph Embedding}

As mentioned in Section \ref{problem_formulation}, skip-gram is a method that generates node features by sampling from a corpus. In this section, we first explain the most commonly used sampling methods for topological graph structures and, through gradient calculations, demonstrate why these methods are unsuitable for navigation in uncertain road networks. Finally, we introduce our sampling method and highlight its advantages.

When generating samples on topological graphs, the most common method is the random walk strategy, \ie~Node2Vec, which is widely used in tasks like node classification and link prediction. Its strength lies in balancing depth-first search (DFS) and breadth-first search (BFS). However, due to the extensive random walks required for each node, it tends to be slow and introduces significant randomness, which adversely affects subsequent training. Most critically, relying on random walks means that the traversability of a particular edge can heavily influence the embedding of preceding nodes, which is detrimental to our task. In the following part, we will briefly explain it through gradient calculations.

First, we introduce the probability from Eq.~\eqref{eqn-2} into Eq.~\eqref{eqn-1}, and then apply the logarithmic transformation:
\begin{equation}\label{eqn-6}
\log \mathbb{P}(w_o \!\mid\! w_c) = \boldsymbol{u}_0^{\top} \boldsymbol{v}_c - \log \left( \sum_{i \in \mathcal{V}} \exp(\boldsymbol{u}_i^{\top} \boldsymbol{v}_c) \right).
\end{equation}
Then we calculate the gradient of the probabilities:

\begin{equation}\label{eqn-7}
\begin{aligned}
\frac{\partial \log \mathbb{P}(w_o \!\mid\! w_c)}{\partial \boldsymbol{v}_c} &= \boldsymbol{u}_o - \frac{\sum_{j \in \mathcal{V}} \exp(\boldsymbol{u}_j^\top \boldsymbol{v}_c) \boldsymbol{u}_j}{\sum_{i \in \mathcal{V}} \exp(\boldsymbol{u}_i^\top \boldsymbol{v}_c)} \\
% &= \boldsymbol{u}_o - \sum_{j \in \mathcal{V}} \left( \frac{\exp(\boldsymbol{u}_j^\top \boldsymbol{v}_c)}{\sum_{i \in \mathcal{V}} \exp(\boldsymbol{u}_i^\top \boldsymbol{v}_c)} \right) \boldsymbol{u}_j \\
&= \boldsymbol{u}_o - \sum_{j \in \mathcal{V}} \mathbb{P}(w_j \!\mid\! w_c) \boldsymbol{u}_j.
\end{aligned}
\end{equation}
According to Eq.~\eqref{eqn-7}, it is evident that the magnitude of the gradient is negatively related to the probability of the context word appearing near the center word. When employing random walks, the frequency at which $n$-hop neighbours' edges appear in the current node's sampling is reduced by a factor of $n$, which is detrimental to our task objectives. 

To address the limitations of traditional Node2Vec for CTP, we implement several adjustments. First, we adjust the random walk mechanism at each node by incorporating edge costs as weights. For each node, we use Dijkstra’s algorithm to compute the shortest paths to all target nodes, forming weighted chains. Finally, we employ skip-gram to generate node embeddings, taking into account the edge weights. This approach reduces randomness, avoids extensive reliance on random walks, and significantly optimizes computation time. 

Subsequently, we enhance the features by concatenating the feature matrix with positional information, including the current position of the ego and target nodes of the others.

\subsection{Attention-based Policy Network}

Different from applying Graph Neural Networks (GNNs) on link prediction tasks \cite{zhang2018link}, we attempt to employ GNNs as a policy network for navigating in changeable road nets. Due to the stochastic nature, traditional GNN algorithms may not be highly suitable. Therefore, we utilize Graph Attention Networks (GATs) that incorporate self-attention mechanisms. This allows our model to dynamically assign different importance weights to the neighbouring nodes, enhancing its representational capability by focusing on more relevant information rather than merely averaging the neighbourhood information.

Unlike GCNs, for $\forall i \in \mathcal{E}$, we calculate the attention coefficient of the neighbours by the following formulation:

\begin{equation}
e_{ij} = \boldsymbol{a}\left(\left[\boldsymbol{W} \boldsymbol{h_i} \!\parallel\! \boldsymbol{W} \boldsymbol{h_j}\right]\right), \quad j \in \mathcal{N}_i,
\end{equation}
where $W$ is the learnable weight matrix used for linear transformations of node features, increasing the vertex features' dimension. The symbol \( \| \) denotes vector concatenation, and the vector \( \boldsymbol{a} \) is used to calculate the attention coefficient, mapping the concatenated high-dimensional features to a scalar value. Then, \(\mathcal{N}_i\) represents the set of neighbouring vertices of node \(v_i\). The attention coefficients for neighbouring nodes are then normalized using the SoftMax function:

\begin{equation}
\alpha_{ij} = \frac{\exp(\text{LeakyReLU}(e_{ij}))}{\sum_{k \in \mathcal{N}_i} \exp(\text{LeakyReLU}(e_{ik}))}.
\end{equation}

\noindent Then, by weighting the features of neighbouring nodes with the normalized attention coefficients \( \alpha_{ij} \), we could obtain the updated node feature representation as:

\begin{equation}
\boldsymbol{h}_i' = \sigma \left( \sum_{j \in \mathcal{N}_i} \alpha_{ij} \boldsymbol{W} \boldsymbol{h}_j \right),
\end{equation}
where \( \sigma \) is a nonlinear activation function. Then by employing multiple attention heads \( K \), the model is able to capture different feature representations:

\begin{equation}
\boldsymbol{h}_i'' = \bigg\|_{k=1}^{K} \sigma \left( \sum_{j \in \mathcal{N}_i} \alpha_{ij}^k \boldsymbol{W}^k \boldsymbol{h}_j \right),
\end{equation}
where \( \big\| \) denotes the concatenation operation across the feature vectors. For any two nodes \( v_i \) and \( v_j \), we calculate the dot product of their feature representations \( \boldsymbol{h}_i'' \) and \( \boldsymbol{h}_j'' \) :

\begin{equation}
\theta_{ij} = \boldsymbol{h}_i''^{\top} \boldsymbol{h}_j''.
\end{equation}
The final step computes the probabilities of selecting actions through the compatibility layer:

\begin{equation}
\pi_{\boldsymbol{\theta}}(a \!\mid\! s) = \frac{\exp(\theta_{ij})}{\sum_{j \in \mathcal{N}_i} \exp(\theta_{ij})}.
\end{equation}
Then during training, we use this probability to guide action selection, while during testing, we directly select the edge corresponding to the maximum value as the strategy.

\subsection{Online Expert-Augmented Policy Gradient}

Our model is trained using the online expert, with the objective function $J_{\boldsymbol{\theta}}$ defined in Eq.~\eqref{eq_1} as the difference between the team's stochastic on-time arrival (SOTA) probability and the information entropy of the road network. Next, we use the centralized controller to compute the prior optimal solution for each state. The model is trained by imitating the decisions of the ``expert". As such, the objective is to minimize the loss $\mathcal{L}$ between the predicted solution and the optimal solution, which essentially corresponds to maximizing the objective function $J_{\boldsymbol{\theta}}$. The gradient of $J_{\boldsymbol{\theta}}$ is defined as follows:

\begin{equation}
\label{eq_15}
\nabla_{\boldsymbol{\theta}} J_{\boldsymbol{\theta}} = \sum\limits_{i \in \mathcal{A}} \lambda_i \nabla_{\boldsymbol{\theta}} J_{\mathbb{P}_i} - \nabla_{\boldsymbol{\theta}} J_{\mathbb{H}}.
\end{equation}

\noindent The next step in the derivation addresses the SOTA probability and information entropy separately. First, we derive the gradient of the SOTA probability using the Variational Policy Gradient (VPG) theorem, defined as follows:

\begin{equation}
\begin{aligned}
\nabla_{\boldsymbol{\theta}} {J}_{\mathbb{P}_i} &= \nabla_{\boldsymbol{\theta}} \mathbb{E}_{\tau \sim \pi_\theta}[\mathbb{P}(G(\tau) \leq T)] \\
% &= \nabla_{\theta} \int_{\tau} \mathbb{P}(G(\tau) \leq T)\pi_\theta(\tau) \, d\tau \\
% &= \int_{\tau} \mathbb{P}(G(\tau) \leq T)\nabla_{\theta}\pi_\theta(\tau) \, d\tau \\
% &= \int_{\tau} \mathbb{P}(G(\tau) \leq T)\nabla_{\theta}\log \pi_\theta(\tau)\pi_\theta(\tau) \, d\tau \\
&= \mathbb{E}_{\tau \sim \pi_{\boldsymbol{\theta}}}[\mathbb{P}(G(\tau) \leq T)\nabla_{\boldsymbol{\theta}}\log \pi_{\boldsymbol{\theta}}(\tau)].
\end{aligned}
\end{equation}
In practice, we adopt the online update method. If our task is treated as an offline task using Monte Carlo sampling, updates will only occur after the trajectory is completed, leading to slow convergence when training on large-scale road networks. Although Monte Carlo sampling has the advantage of being an unbiased estimator, it is not well-suited for continuous tasks. To address this, we use the sigmoid function in combination with the agent's remaining time with the cost and variance of the shortest remaining path as the metric for on-time arrival. Consequently, we derive the following:

\begin{equation}
\label{eq_17}
\begin{aligned}
\nabla_{\boldsymbol{\theta}} {J}_{\mathbb{P}_i} &= \mathbb{E}_{\tau \sim \pi_{\boldsymbol{\theta}}}[\mathbb{P}(G(\tau) \leq T)\nabla_{\boldsymbol{\theta}}\log \pi_\theta(\tau)] \\
&\approx \frac{1}{M} \sum_{j=1}^{M} \mathbb{S}\{G(\tau^{(j)}) \leq T\}\nabla_{\boldsymbol{\theta}}\log \pi_{\boldsymbol{\theta}}(\tau^{(j)}),
% &= \frac{1}{M} \sum_{j=1}^{M} \mathbb{S}\{G(\tau^{(j)}) \leq T\} \sum_{k=0}^{H_j} \nabla_{\theta} \log \pi_\theta(a \mid s_j),
\end{aligned}
\end{equation}
where \(M\) represents the total timesteps for the corresponding robot, while $\mathbb{S}$ represents the sigmoid function, used to calculate the robot's SOTA probability. Next, we calculate the gradient of the information entropy, where its gradient with respect to the policy can be expressed as follows:

\begin{equation}
\nabla_{\boldsymbol{\theta}} J_{\mathbb{H}} = \mathbb{E}_{\tau \sim \pi_{\boldsymbol{\theta}}} \left[ \nabla_{\boldsymbol{\theta}} \log \pi_{\boldsymbol{\theta}}(\tau) \cdot \Delta \mathbb{H} (\tau) \right], 
\end{equation}
where $\mathbb{H}$ is defined in Eq.~\eqref{eqn-4}, $\Delta \mathbb{H}(\tau)$ represents the change in the information entropy of the overall graph after the agent follows the trajectory according to the policy:

\begin{equation}
\label{eq_19}
\Delta\mathbb{H} = \sum_{e \in {\mathcal{E}}_{\text{explored}}(\tau)} \left( p(e) \log p(e) \right),
\end{equation}
where $\mathcal{E}_{\text{explored}}(\tau)$ represents the set of edges explored by the robot in trajectory $\tau \sim \pi_{\boldsymbol{\theta}}$. The information entropy is also updated online, resulting in the following:

\begin{equation}
\label{eq_20}
\begin{aligned}
\nabla_{\boldsymbol{\theta}} J_{\mathbb{H}} &= \mathbb{E}_{\tau \sim \pi_{\boldsymbol{\theta}}} [ \Delta \mathbb{H} \cdot \nabla_{\boldsymbol{\theta}} \log \pi_{\boldsymbol{\theta}}(\tau) ]\\
&\approx \frac{1}{M} \sum_{j=1}^{M} \Delta \mathbb{H} \cdot\nabla_{\boldsymbol{\theta}}\log \pi_{\boldsymbol{\theta}}(\tau^{(j)}).
\end{aligned}
\end{equation}

\noindent
By combining Eq.~\eqref{eq_15},  Eq.~\eqref{eq_17}, Eq.~\eqref{eq_19}, and Eq.~\eqref{eq_20},  we ultimately obtain the gradient of our objective $J_{\theta}$:
\begin{equation*} 
\begin{aligned}
\nabla_{\boldsymbol{\theta}} J_{\boldsymbol{\theta}} &= \sum_{i \in A} \lambda_i \left( \frac{1}{M} \sum_{j=1}^{M} \mathbb{S}\{G(\tau^{(j)}) \leq T\} \cdot \nabla_\theta \log \pi_{\boldsymbol{\theta}}(\tau^{(j)}) \right)
\end{aligned}
\end{equation*}
\begin{equation}\label{eq_21} % 这里保留编号为21
\begin{aligned}
& \quad - \frac{1}{M} \sum_{j=1}^{M} \Delta \mathbb{H} \cdot \nabla_{\boldsymbol{\theta}} \log \pi_{\theta}(\tau^{(j)}).
\end{aligned}
\end{equation}
The subsequent gradient computation will proceed via the chain rule, propagating and updating accordingly. As this process falls under the standard backpropagation algorithm, the detailed derivation will not be elaborated in this letter. The main process of MARVEL is shown as Algorithm~\ref{algorithm_1}.

\begin{algorithm}\label{algorithm_1}
    \caption{Expert-Augmented Policy Gradient}
    \KwIn{(1)~Graph $\mathcal{G}=(\mathcal{V},\mathcal{E})$;(2)~learning rate $\alpha_{\boldsymbol{\theta}}$; (3)~current vertex $v_i$; (4)~importance weight $\lambda_i$; (5)~destination $d_i$;  (6)~time budget $T_i$; 
   ($i\in\mathcal{R}$). }
   
    \KwOut{Robot Action $a_t = v'_i$. }
    
    \While{$v_i \neq d_i$ and $t_{spent} < T_i$}{
        Execute policy $\pi_{\boldsymbol{\theta}}$ with action $a_t = \text{SoftMax}(\pi_{\boldsymbol{\theta}})$;
        
        Calculate $\nabla_{\boldsymbol{\theta}} J_{\boldsymbol{\theta}}$ according to Eq.~\eqref{eq_21};
        
        Update the parameter vector of policy $\pi$  as:
        $\boldsymbol{\theta} = \boldsymbol{\theta} + \alpha_{\boldsymbol{\theta}} \nabla_{\boldsymbol{\theta}}{J}_{\boldsymbol{\theta}};$
    } 

    \Return Learned parameter vector $\boldsymbol{\theta}$.
\end{algorithm}
\vspace{-5 pt}

\section{Simulation Results and Analysis}
\label{Simulation Results and Analysis}
In this section, we first introduce the evaluation metrics and then describe our experimental setup. Following this, we evaluate MARVEL on a representative road network and then expand the evaluation to broader networks for generalizability. We also provide a comparison with baseline algorithms.
\vspace{-5pt}

\subsection{Metrics and Setup}
\subsubsection{Metrics}
The primary evaluation metric is the team’s probability of on-time arrival, calculated as a weighted sum based on each robot’s task priority. For each robot with an origin-destination (OD) pair, we estimate the on-time success rate by running 10,000 simulations based on actual travel time.

\textit{Travel Cost Sampling:} Real travel costs are sampled from a Gaussian distribution with mean \(\mu\) and standard deviation \(\sigma\), consistent with realistic network conditions. To account for minimal travel times, we truncate samples with rejection, ensuring alignment with real-world road network constraints.

\subsubsection{Dataset}
To verify the effectiveness of MARVEL, we conduct benchmark experiments by comparing its performance against baseline algorithms on a series of canonical transportation networks, namely Sioux Falls Network, Friedrichshain,
Anaheim, and Winnipeg, publicly available in \cite{Bar-Gera2021}. 

\subsubsection{Setup}
The training utilizes the Adam optimizer with a learning rate of \(10^{-3}\) and a decay factor of 0.95 of 100 epochs. The feature dimension was set to 128, with an 8-head attention mechanism. Training is conducted on a setup with Intel Core i9-13900KF CPU and NVIDIA 4090 GPU. The source code is available at \url{https://github.com/YUJ0E/MARVEL}.

% \begin{table}[h]
% \centering
% \caption{Canonical Transportation Networks}
% \label{table_1}
% \begin{tabular}{lcc}
% \hline
% \textbf{Test networks} & \textbf{\# of edges $(m)$} & \textbf{\# of nodes $(n)$} \\
% \hline
% Sioux Falls Network & 76 & 24 \\
% Friedrichshain & 523 & 224 \\
% Anaheim & 914 & 416 \\
% Winnipeg & 2836 & 1052 \\
% \hline
% \end{tabular}
% \end{table}

\subsection{Baselines}
As previously mentioned, existing CTP solutions are based on single-agent algorithms and the Multi-agent Path Finding algorithms are predominantly designed for grid-based maps. For our benchmark experiments, each agent is thus evaluated using the single-agent algorithms with predetermined traversal probabilities for unknown edges. For comparison, we selected three representative state-of-the-art RSP algorithms as baselines: (1) FMA\cite{guo2022navigation}: Fourth-Moment Approach, (2) PQL\cite{cao2020using}: Practical Q-learning, and (3) DOT\cite{prakash2020algorithms}: Decreasing Order-of-Time. Additionally, we included three canonical CTP solution algorithms: (1) \(\pi(\tau)\)\cite{chen2017most}: Reliable Path Search Algorithm, (2) RAO*\cite{guo2019robust}: Robust Risk-Aware Path Planning, and (3) DDP\cite{guo2022dual}: Dual Dynamic Programming. These serve as the baseline algorithms as the comparison basis.

\subsection{Performance Evaluation with the Simple Network}
\begin{figure}[htbp]
    \centering
    % 第一行的两个图
    \subfigure[Robot A Success Rate]{
        \includegraphics[width=0.45\linewidth, keepaspectratio]{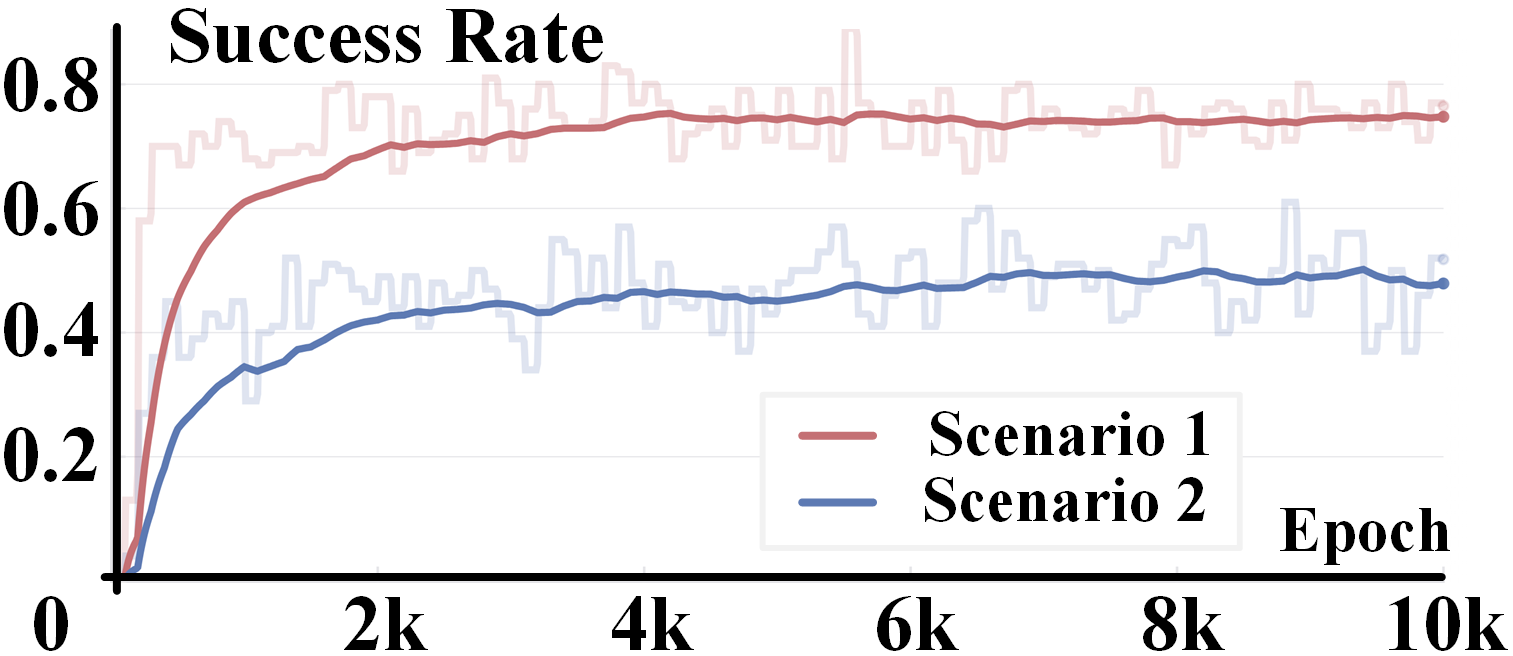}
        \label{robotAsr}
    }
    \hfill
    \subfigure[Robot B Success Rate]{
        \includegraphics[width=0.45\linewidth, keepaspectratio]{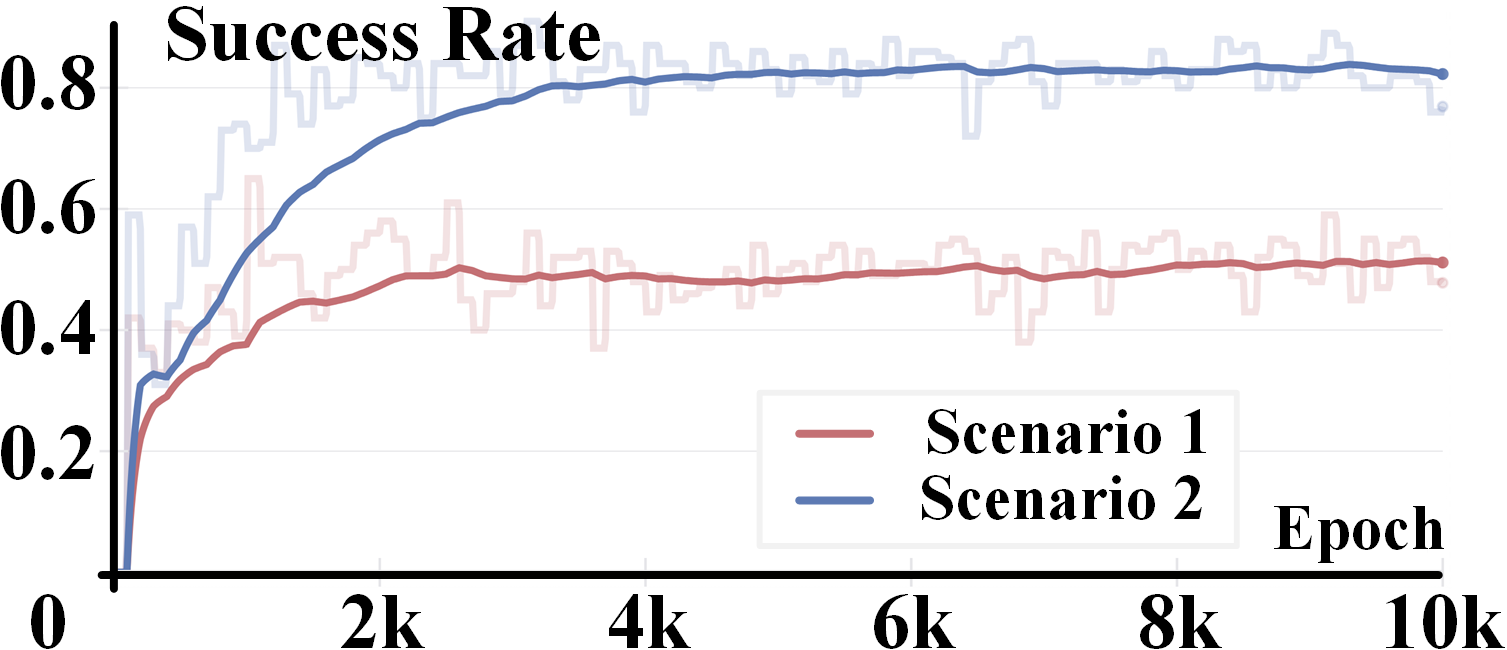}
        \label{robotBsr}
    }
    % \hfill
    % \subfigure[Differences on Robot A]{
    %     \includegraphics[width=0.15\linewidth, keepaspectratio]{Figures/zoomin.pdf}
    %     \label{Zoom}
    % }
    \caption{The SOTA probability performance comparison of different scenarios.}
    \label{success_rate}
\end{figure}

Initially, we validated MARVEL on the illustrative network in Fig.~\ref{fig_1}. We configured two scenarios: in Scenario 1, Robot A has a higher task priority than Robot B ($\lambda = [0.7, 0.3]^{\top}$). In Scenario 2, Robot A has a lower priority ($\lambda = [0.3, 0.7]^{\top}$). The final results of the two robots are displayed in Fig~\ref{success_rate}.

% \begin{table}[h]
% \centering
% \caption{The SOTA performance comparison of different scenarios}
% \label{table_sota_comparison}
% \begin{tabular}{lcccc}
% \toprule
% \multirow{2}{*}{} & \multicolumn{2}{c}{\textbf{Robot A}} & \multicolumn{2}{c}{\textbf{Robot B}} \\
% \cmidrule(lr){2-3} \cmidrule(lr){4-5}
%                   & Weight & SOTA Probability & Weight & SOTA Probability \\
% \midrule
% Scenario 1        & 0.3    & 0.42     & 0.7    & \textbf{0.72}      \\
% Scenario 2        & 0.5    & \textbf{0.67}     & 0.5    & 0.61      \\
% \bottomrule
% \end{tabular}
% \end{table}

Based on the results, in Scenario 1, where Robot A has a higher priority, Robot A prioritizes its own arrival, while Robot B takes the risk of delay if the edge from 13 to 4 becomes impassable. Conversely in Scenario 2, Robot A has a lower priority. As a result, it sacrifices its arrival probability to conduct an exploratory action to node 13 instead of node 11 which benefits the team and Robot B's on-time arrival probability is increased as it adapts its route based on observations made by Robot A. Overall, MARVEL guides robots to make team-optimal decisions based on assigned priority weights.

\subsection{Performance Comparison with Baseline Algorithms}
 In this subsection, we evaluate the performance of MARVEL with advanced baseline algorithms. For Friedrichshain, Anaheim, and Winnipeg, we randomly generated 50 OD pairs with stochastic weights, while for the Sioux Falls Network (SFN), we generated 10 such OD pairs, due to its smaller scale. Please note that MARVEL allocates weights based on task priority, and we set the time budget as a multiple of the least expected time $t_\text{LET}$. Thus, we set the team on-time arrival probability for low-priority OD pairs using $1.2 \times t_\text{LET}$ as the corresponding time budget. For high-priority OD pairs, based on the study in \cite{gao2021gp4}, we assess each OD pair with three use cases: $T = 0.95 \times t_\text{LET}$ (tight schedule), $T = 1.0 \times t_\text{LET}$ (exact schedule), and $T = 1.05 \times t_\text{LET}$ (relaxed schedule).
\begin{figure*}[htbp]
    \centering
    % 第一行的两个图
    \subfigure[Sioux Falls Network]{
        \includegraphics[width=0.23\linewidth, keepaspectratio]{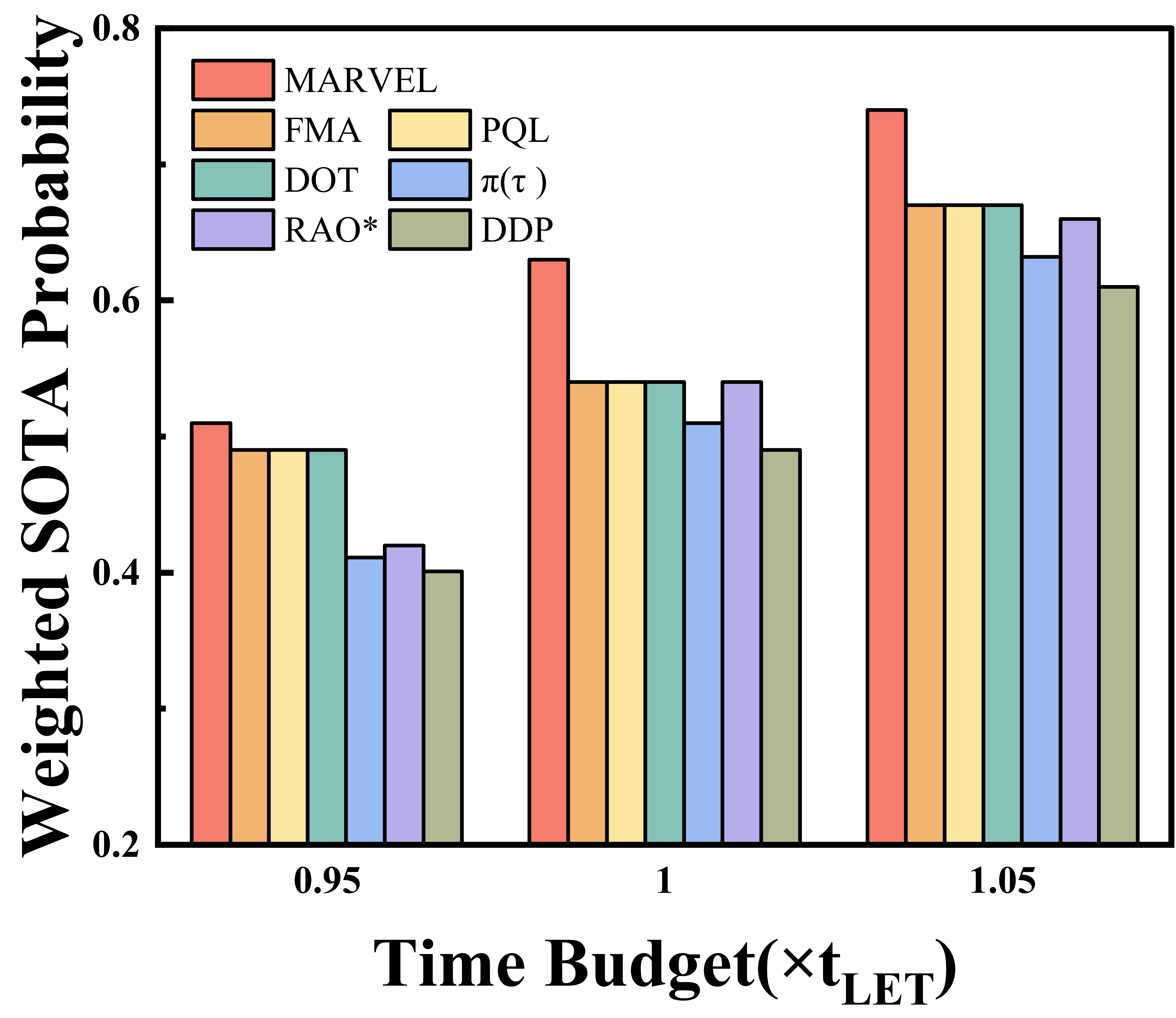}
        \label{fig:sfn-prob}
    }
    \hfill
    \subfigure[Friedrichshain]{
        \includegraphics[width=0.23\linewidth, keepaspectratio]{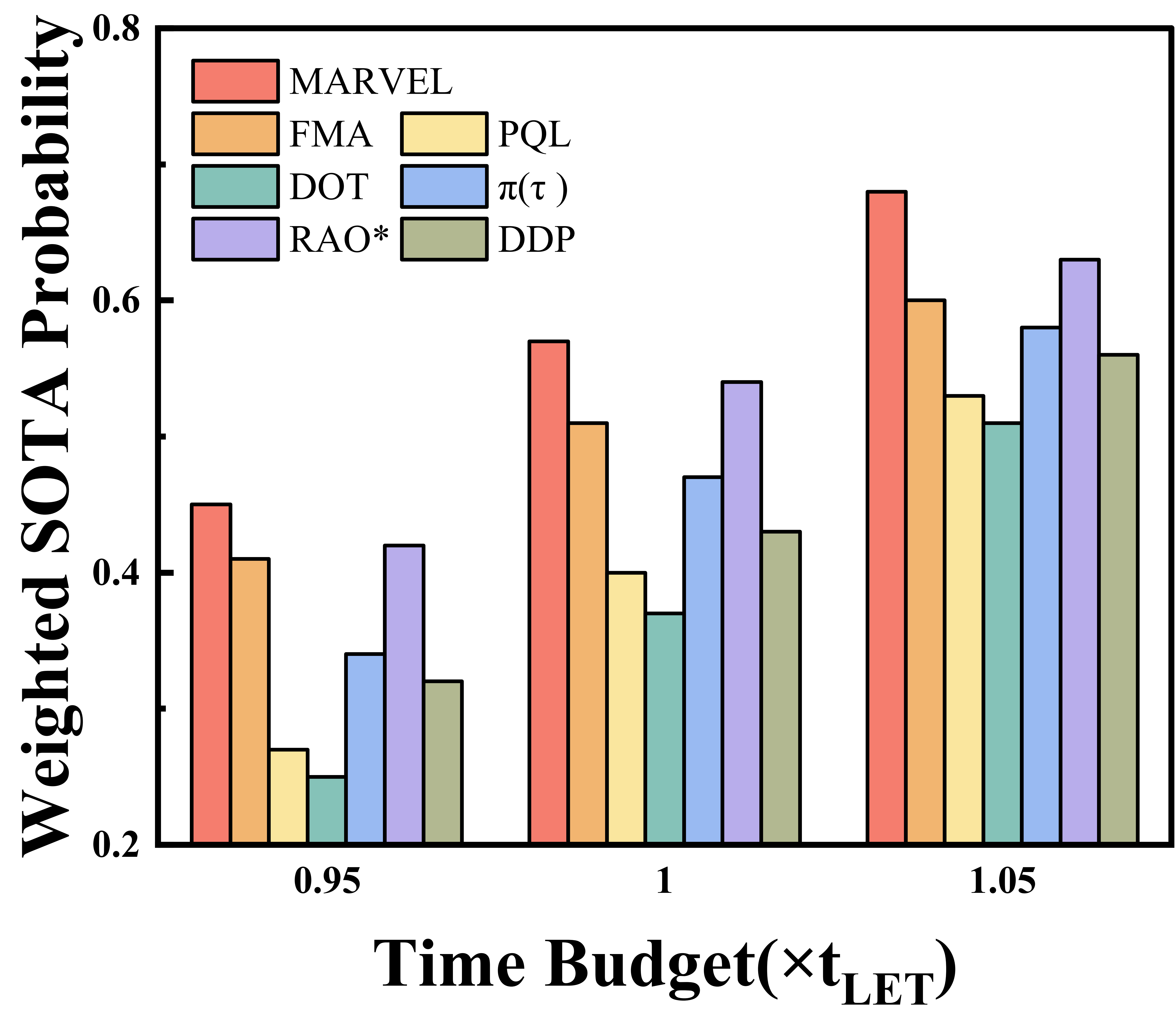}
        \label{fig:fried-prob}
    }
    \hfill
    \subfigure[Anaheim]{
        \includegraphics[width=0.23\linewidth, keepaspectratio]{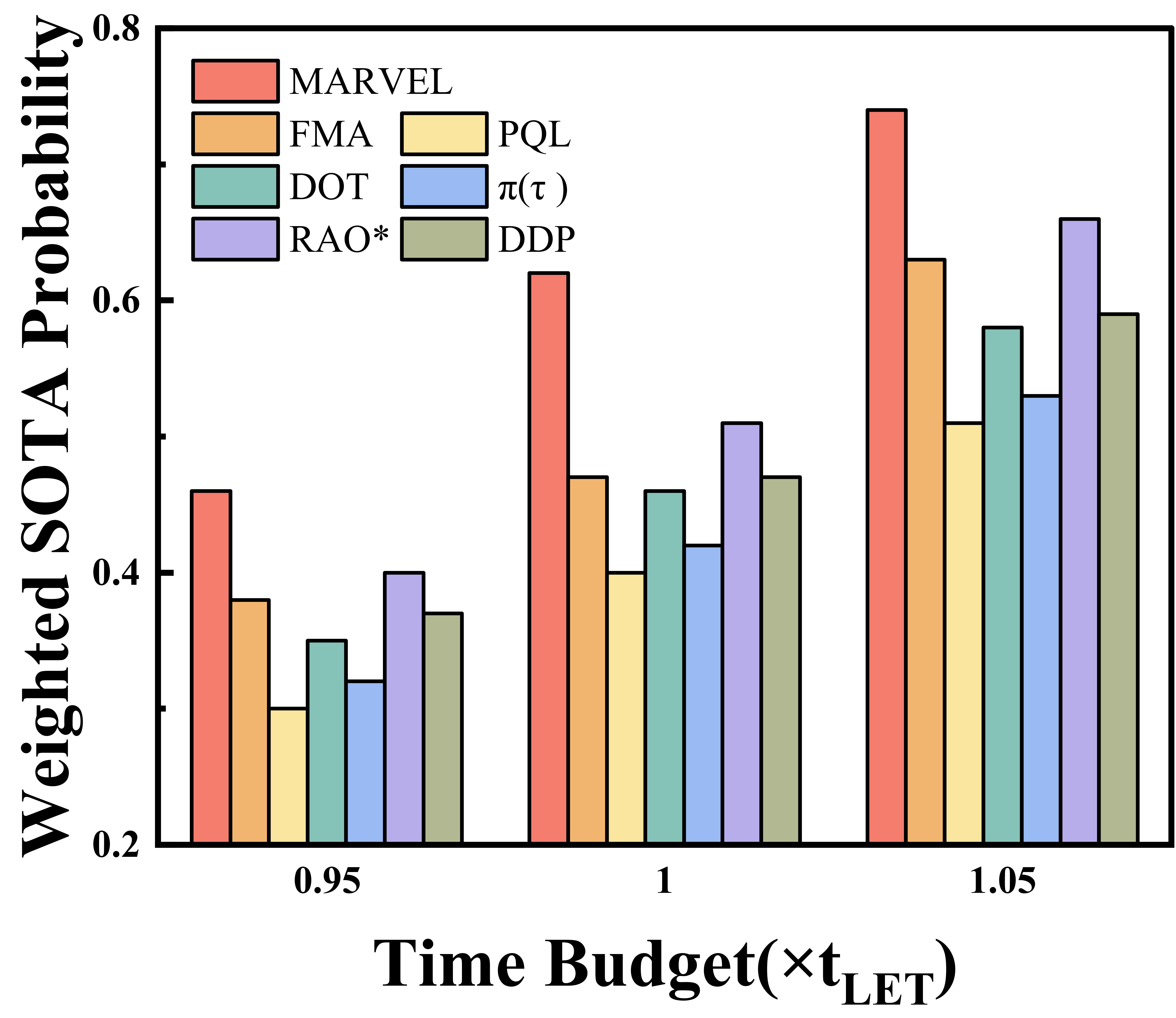}
        \label{fig:anaheim-prob}
    }
    \hfill
    \subfigure[Winnipeg]{
        \includegraphics[width=0.23\linewidth, keepaspectratio]{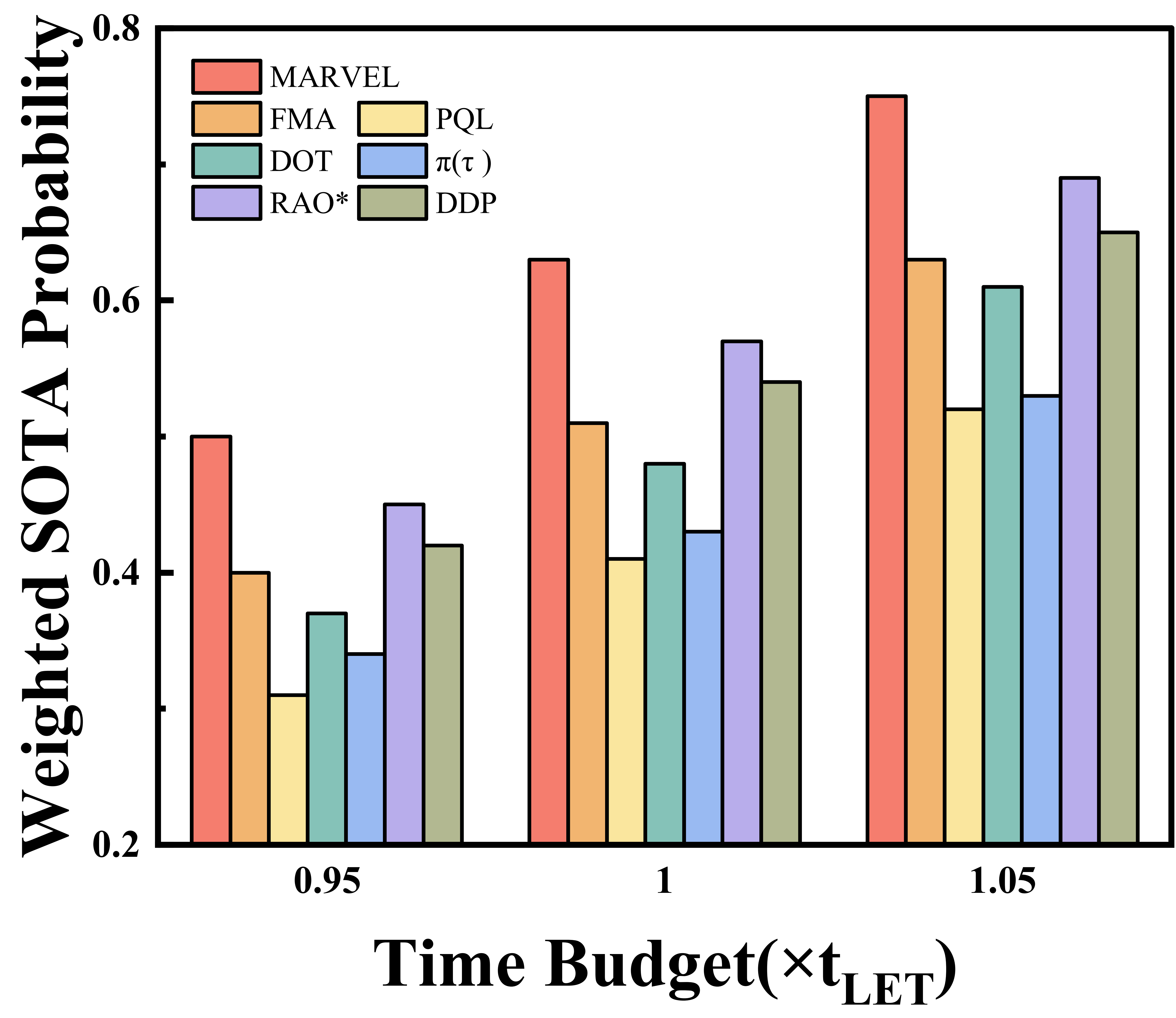}
        \label{fig:winnipeg-prob}
    }
    
    \caption{The SOTA probability performance comparison with the baseline algorithms on different canonical transportation networks.}
    \label{fig:perf-comparison-four}
\end{figure*}

Fig.~\ref{fig:perf-comparison-four} compares the SOTA probability performance achieved by our MARVEL versus the baseline algorithms. From the results, we can observe that: (1) MARVEL consistently outperforms other state-of-the-art algorithms and CTP solutions, further highlighting the significant advantages of collective intelligence across different time budgets and varying transportation network scales. (2) The SOTA probability for the tight budget scenario is considerably lower than in the other cases. This discrepancy is due not only to the highly constrained time budget but also to our sampling method, which employs a truncated sampling strategy with rejection to ensure the minimum value relative to the overall average.

\subsection{Ablation Study}
We conduct an ablation study to evaluate the effectiveness of our proposed algorithm. The ablation experiments include comparisons with the canonical Node2Vec, removal of the self-attention mechanism within the network, and exclusion of information entropy and cross-entropy loss components. The study is conducted based on the exact schedule of the SFN.  The decreases and increases in parentheses are relative to the initial data proportions from MARVEL.

\begin{table}[htbp]
    \centering
    \caption{Ablation Performance of MARVEL.}
    \label{ablation}
    \begin{tabular}{lcc}
        \toprule
        \textbf{Method} & \textbf{SOTA Probability} & \textbf{Converge Epochs} \\
        \midrule
        \textbf{MARVEL} & \textbf{63\%} & \textbf{4,000} \\
        Canonical Node2Vec & Not Convergent & Not convergent \\
        No Self-attention & 40\% \,(37\%$\downarrow$) & 5500 (37.5\%$\uparrow$) \\
        No Information Entropy & 57\% \,(10\%$\downarrow$) & 4000 \\
        No Cross Entropy Loss & 62\% & 7000 (75\%$\uparrow$) \\
        \bottomrule
    \end{tabular}
\end{table}

Based on Table~\ref{ablation}, when applying canonical Node2Vec to preprocess graph features, the results do not converge, indicating that Node2Vec is not suitable for stochastic networks. Additionally, both the self-attention mechanism and information entropy have a significant impact on SOTA probability. In particular, ignoring attention results in the algorithm's inability to accurately judge whether the passability of the unknown edges affects the current robot. Information entropy serves as an important criterion for determining whether low-priority robots should engage in exploration. As for cross-entropy loss, although it does not have a substantial impact on the results, it significantly reduces the convergence speed.

\section{Physical Experiment Validation}
\label{Deployment and Experiment Results}
\begin{figure}[htbp]
    \centering
    % 第一行的两个图
    \subfigure[Physical maze setup (Scenario A)]{
        \includegraphics[width=0.52\linewidth, keepaspectratio]{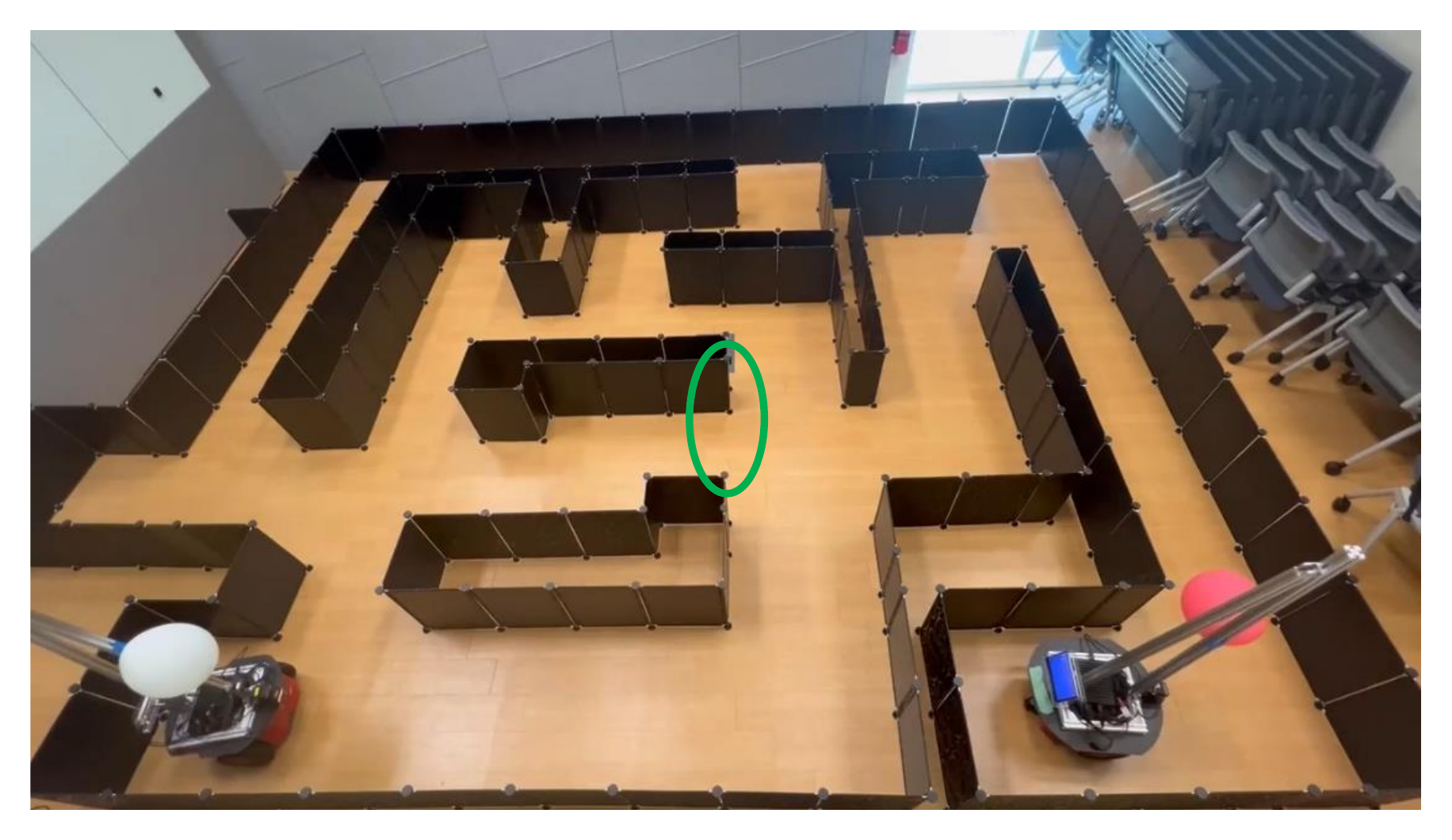}
        \label{passA}
    }
    \hfill
    \subfigure[RViz Map (Scenario A)]{
        \includegraphics[width=0.39\linewidth, keepaspectratio]{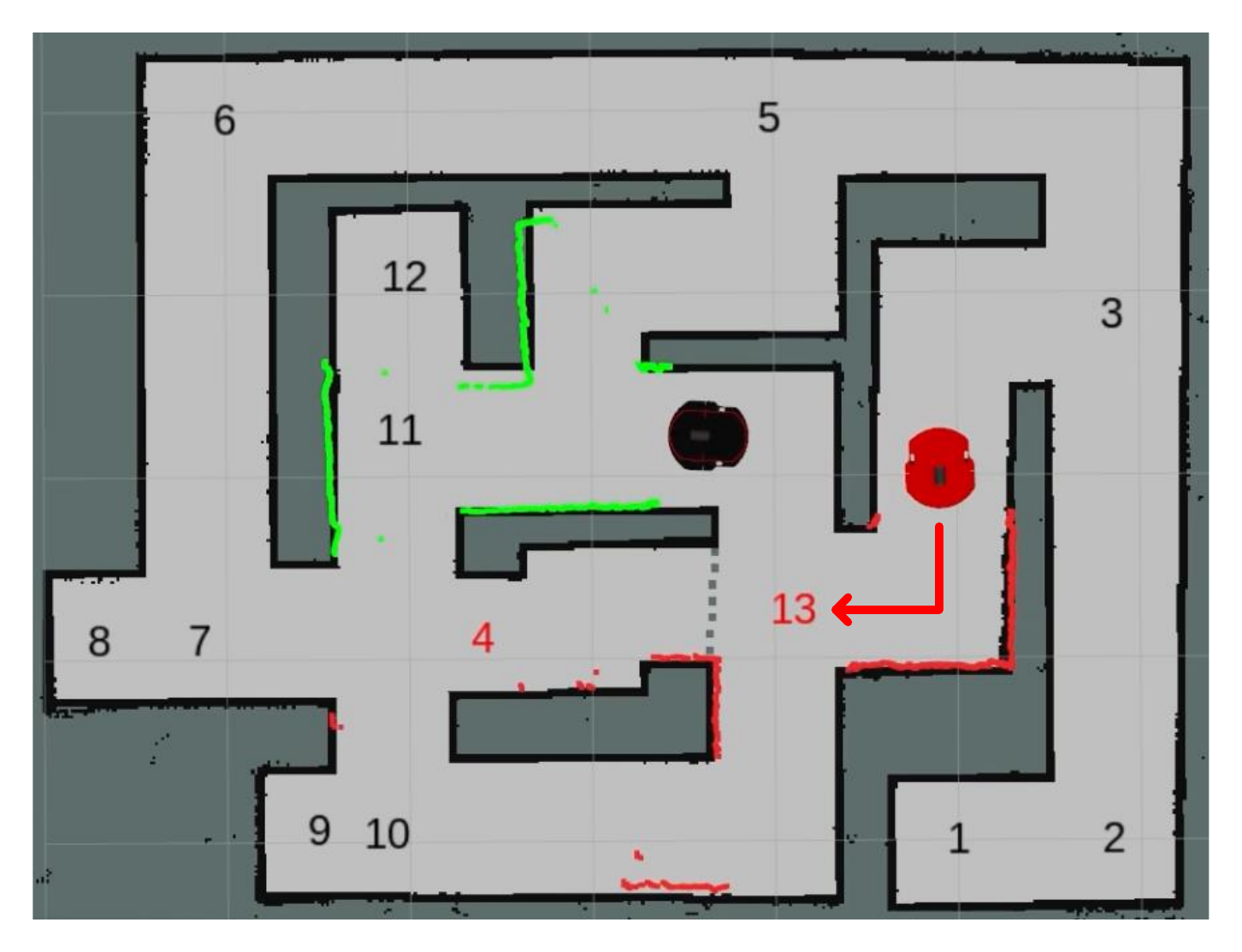}
        \label{passgazebo}
    }
    \hfill
    \subfigure[Physical maze setup (Scenario B)]{
        \includegraphics[width=0.52\linewidth, keepaspectratio]{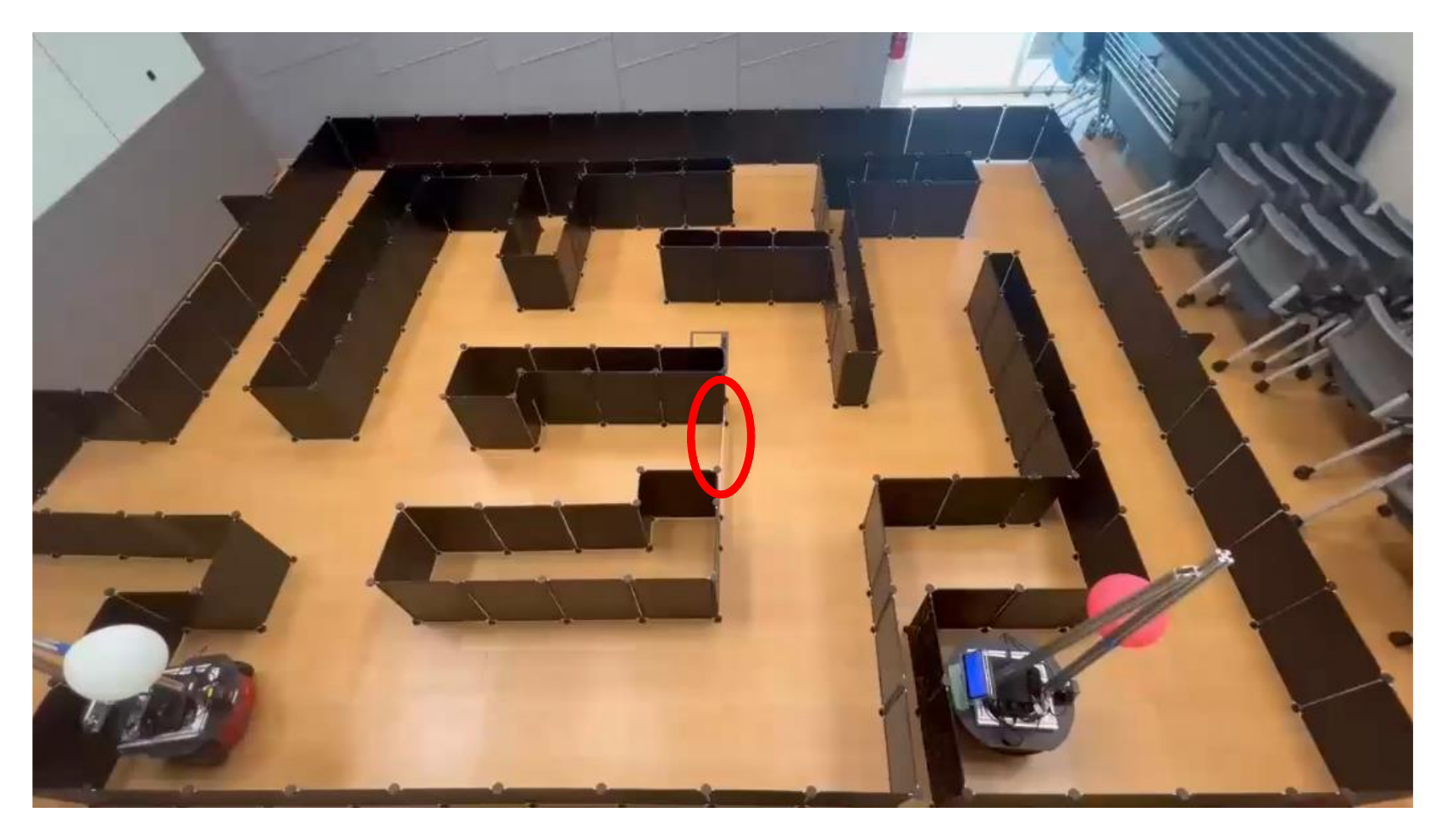}
        \label{passb}
    }
    \hfill
    \subfigure[RViz Map (Scenario B)]{
        \includegraphics[width=0.38\linewidth, keepaspectratio]{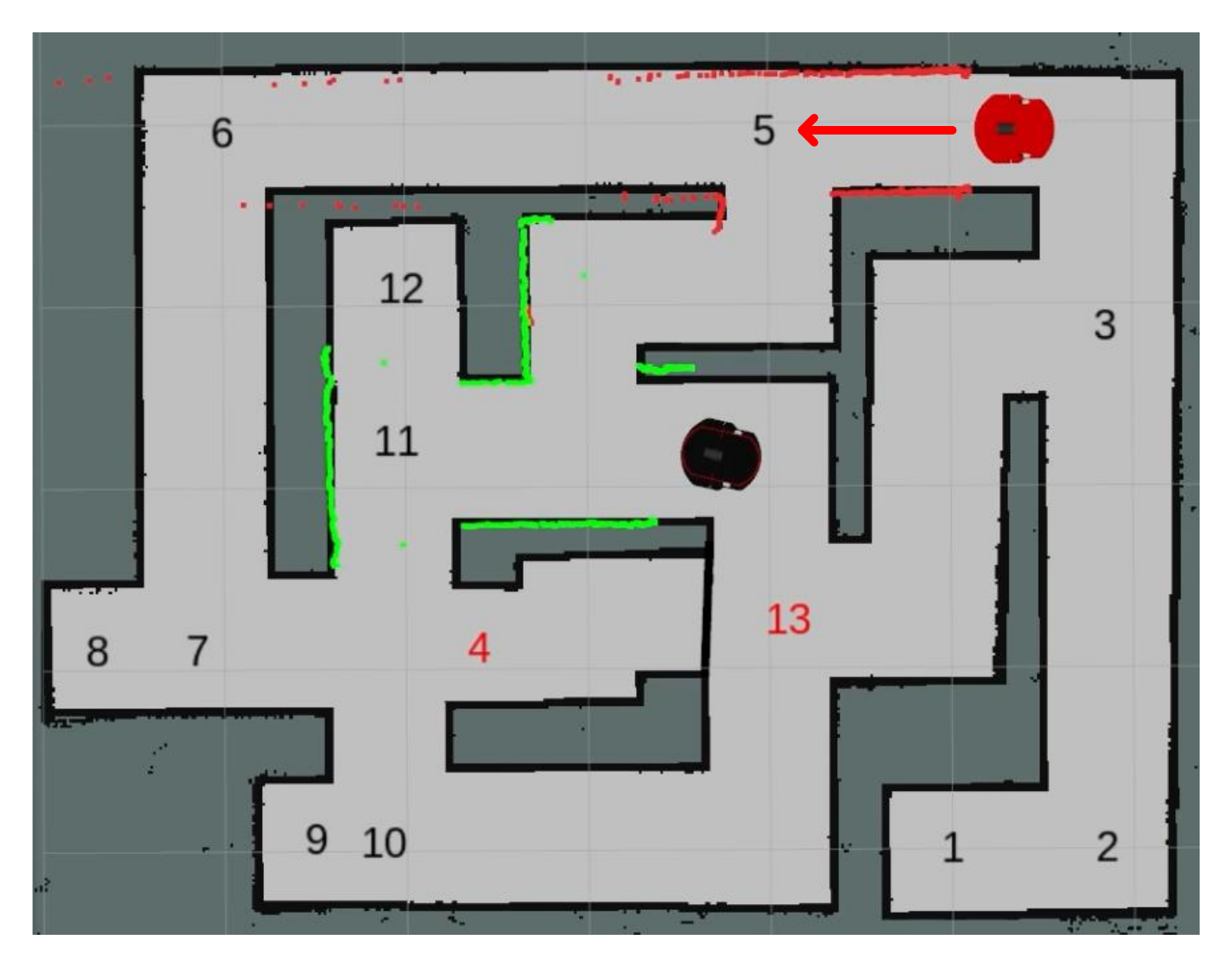}
        \label{passbgazebo}
    }
    \caption{Constructed physical environments and Rviz Virtual Maps.}
    \label{fig:realworld}
\end{figure}

In this section, we deploy MARVEL on a real multi-robot system and set up a physical environment to conduct real-world experiments, assessing its performance under dynamic and uncertain conditions. The experimental platform is the Pioneer 3-DX robot, a robust differential-drive robot equipped with sensors for localization and navigation. The maze environment is constructed based on the topological map shown in Fig.~\ref{fig_1}, mirroring the nodes and edges in the original graph to simulate real-world navigation scenarios. We designed two sets of experiments: in Scenario A (illustrated in Fig.~\ref{passA} and Fig.~\ref{passgazebo}), edge 13-4 is passable, allowing for an optimal route; in Scenario B (depicted in Fig.~\ref{passb} and Fig.~\ref{passbgazebo}), edge 13-4 is impassable, requiring the robots to adjust and take an alternative route. These scenarios allow us to evaluate MARVEL’s adaptability and decision-making effectiveness in environments with uncertain traversal paths.

After training, we test the model in a physical environment. Robot A (marked with a white balloon) detours to node 13 to observe the passability of edge 13-4, which informs Robot B’s (marked with a red balloon) route. If the edge is passable, Robot B heads to node 13 to take the shortest path. Otherwise, it reroutes to node 5, adapting its route based on Robot A’s observations. The demonstration video is uploaded with the manuscript and also can be found in the GitHub repository.

\section{Conclusion and Future Work}
\label{Conclusion and Future Work}

In this paper, we propose a novel multi-agent reinforcement learning algorithm that integrates GAT to enhance navigation in varying topological networks with unknown traversal probabilities. We refine the graph embedding by embedding probabilities and O-D information into the feature matrix, increasing the adaptability to the dynamical network. Through the entropy-based online expert, MARVEL balances exploration and exploitation, promoting cooperation to maximize the team's SOTA probability. Benchmark experiments demonstrate MARVEL's superior performance over the baseline algorithms, even without prior edge traversal data, underscoring the benefits of collective intelligence in dynamic environments.

In future work, we aim to enhance MARVEL's generalization and computational efficiency. Specifically, we will address the inherent complexity of topological networks and the challenges of uncertain edge information, which may limit adaptability across different environments. We also plan to conduct experiments in more diverse and complex environments beyond the constructed maze setup to further validate the applicability of the proposed approach. Additionally, improving the computational efficiency will increase the responsiveness of the proposed approach which is useful for real-time applications in large-scale road networks.

% \section*{Acknowledgement}
% This work was supported in part by the Robotics Horizontal Technology Coordinating Office, Agency for Science, Technology and Research, Singapore, under Grant C221518004.
% \url{http://www.latex-community.org/}
%{\appendices
%\section*{Proof of the First Zonklar Equation}
%Appendix one text goes here.
% You can choose not to have a title for an appendix if you want by leaving the argument blank
%\section*{Proof of the Second Zonklar Equation}
%Appendix two text goes here.}

% \section{References Section}
% You can use a bibliography generated by BibTeX as a .bbl file.
%  BibTeX documentation can be easily obtained at:
%  http://mirror.ctan.org/biblio/bibtex/contrib/doc/
%  The IEEEtran BibTeX style support page is:
%  http://www.michaelshell.org/tex/ieeetran/bibtex/
 
 % argument is your BibTeX string definitions and bibliography database(s)
%\bibliography{IEEEabrv,../bib/paper}
%
% \section{References}
% You can manually copy in the resultant .bbl file and set second argument of $\backslash${\tt{begin}} to the number of references
%  (used to reserve space for the reference number labels box).

\bibliographystyle{IEEEtran}

\bibliography{IEEEabrv,citations}

% Generated by IEEEtran.bst, version: 1.14 (2015/08/26)
\begin{thebibliography}{10}
\providecommand{\url}[1]{#1}
\csname url@samestyle\endcsname
\providecommand{\newblock}{\relax}
\providecommand{\bibinfo}[2]{#2}
\providecommand{\BIBentrySTDinterwordspacing}{\spaceskip=0pt\relax}
\providecommand{\BIBentryALTinterwordstretchfactor}{4}
\providecommand{\BIBentryALTinterwordspacing}{\spaceskip=\fontdimen2\font plus
\BIBentryALTinterwordstretchfactor\fontdimen3\font minus \fontdimen4\font\relax}
\providecommand{\BIBforeignlanguage}[2]{{%
\expandafter\ifx\csname l@#1\endcsname\relax
\typeout{** WARNING: IEEEtran.bst: No hyphenation pattern has been}%
\typeout{** loaded for the language `#1'. Using the pattern for}%
\typeout{** the default language instead.}%
\else
\language=\csname l@#1\endcsname
\fi
#2}}
\providecommand{\BIBdecl}{\relax}
\BIBdecl

\bibitem{guo2019robust}
H.~Guo and T.~D. Barfoot, ``The robust {Canadian} traveller problem applied to robot routing,'' in \emph{2019 International Conference on Robotics and Automation (ICRA)}.\hskip 1em plus 0.5em minus 0.4em\relax IEEE, 2019, pp. 5523--5529.

\bibitem{cao2020using}
Z.~Cao, H.~Guo, W.~Song, K.~Gao, Z.~Chen, L.~Zhang, and X.~Zhang, ``Using reinforcement learning to minimize the probability of delay occurrence in transportation,'' \emph{IEEE Transactions on Vehicular Technology}, vol.~69, no.~3, pp. 2424--2436, 2020.

\bibitem{wang2022multi}
H.~Wang and W.~Chen, ``Multi-robot path planning with due times,'' \emph{IEEE Robotics and Automation Letters}, vol.~7, no.~2, pp. 4829--4836, 2022.

\bibitem{yang2024attention}
Y.~Yang, M.~Fan, C.~He, J.~Wang, H.~Huang, and G.~Sartoretti, ``Attention-based priority learning for limited time multi-agent path finding,'' in \emph{Proceedings of the 23rd International Conference on Autonomous Agents and Multiagent Systems}, 2024, pp. 1993--2001.

\bibitem{cao2016multiagent}
Z.~Cao, H.~Guo, J.~Zhang, and U.~Fastenrath, ``Multiagent-based route guidance for increasing the chance of arrival on time,'' in \emph{Proceedings of the AAAI Conference on Artificial Intelligence}, vol.~30, no.~1, 2016.

\bibitem{ma2017multi}
H.~Ma, T.~S. Kumar, and S.~Koenig, ``Multi-agent path finding with delay probabilities,'' in \emph{Proceedings of the AAAI Conference on Artificial Intelligence}, vol.~31, no.~1, 2017.

\bibitem{shiri2017online}
D.~Shiri and F.~S. Salman, ``On the online multi-agent o--d k-{Canadian} traveller problem,'' \emph{Journal of Combinatorial Optimization}, vol.~34, pp. 453--461, 2017.

\bibitem{papadimitriou1991shortest}
C.~H. Papadimitriou and M.~Yannakakis, ``Shortest paths without a map,'' \emph{Theoretical Computer Science}, vol.~84, no.~1, pp. 127--150, 1991.

\bibitem{guzzi2019impact}
J.~Guzzi, R.~O. Chavez-Garcia, L.~M. Gambardella, and A.~Giusti, ``On the impact of uncertainty for path planning,'' in \emph{2019 International Conference on Robotics and Automation (ICRA)}.\hskip 1em plus 0.5em minus 0.4em\relax IEEE, 2019, pp. 5929--5935.

\bibitem{liao2015generalized}
C.-S. Liao and Y.~Huang, ``Generalized {Canadian} traveller problems,'' \emph{Journal of Combinatorial Optimization}, vol.~29, pp. 701--712, 2015.

\bibitem{nikolova2008route}
E.~Nikolova and D.~R. Karger, ``Route planning under uncertainty: The {Canadian} traveller problem.'' in \emph{AAAI}, 2008, pp. 969--974.

\bibitem{bender2015optimal}
M.~Bender and S.~Westphal, ``An optimal randomized online algorithm for the k-{Canadian} traveller problem on node-disjoint paths,'' \emph{Journal of Combinatorial Optimization}, vol.~30, no.~1, pp. 87--96, 2015.

\bibitem{su2009recoverable}
B.~Su and X.~LAN, ``The recoverable {Canadian} traveller problem based on limited provision information,'' \emph{Syst. Eng.}, vol.~9, pp. 102--107, 2009.

\bibitem{liao2014covering}
C.-S. Liao and Y.~Huang, ``The covering {Canadian} traveller problem,'' \emph{Theoretical Computer Science}, vol. 530, pp. 80--88, 2014.

\bibitem{okoso2021network}
A.~Okoso, B.~Okumura, K.~Otaki, and T.~Nishi, ``Network-flow-problem-based approach to multi-agent path finding for connected autonomous vehicles,'' in \emph{2021 IEEE International Intelligent Transportation Systems Conference (ITSC)}.\hskip 1em plus 0.5em minus 0.4em\relax IEEE, 2021, pp. 1946--1953.

\bibitem{ma2021distributed}
Z.~Ma, Y.~Luo, and H.~Ma, ``Distributed heuristic multi-agent path finding with communication,'' in \emph{2021 IEEE International Conference on Robotics and Automation (ICRA)}.\hskip 1em plus 0.5em minus 0.4em\relax IEEE, 2021, pp. 8699--8705.

\bibitem{he2024social}
C.~He, T.~Duhan, P.~Tulsyan, P.~Kim, and G.~Sartoretti, ``Social behavior as a key to learning-based multi-agent pathfinding dilemmas,'' \emph{arXiv preprint arXiv:2408.03063}, 2024.

\bibitem{ma2017feasibility}
H.~Ma, J.~Yang, L.~Cohen, T.~Kumar, and S.~Koenig, ``Feasibility study: Moving non-homogeneous teams in congested video game environments,'' in \emph{Proceedings of the AAAI Conference on Artificial Intelligence and Interactive Digital Entertainment}, vol.~13, no.~1, 2017, pp. 270--272.

\bibitem{alkazzi2024comprehensive}
J.-M. Alkazzi and K.~Okumura, ``A comprehensive review on leveraging machine learning for multi-agent path finding,'' \emph{IEEE Access}, 2024.

\bibitem{he2024alpha}
C.~He, T.~Yang, T.~Duhan, Y.~Wang, and G.~Sartoretti, ``Alpha: Attention-based long-horizon pathfinding in highly-structured areas,'' in \emph{2024 IEEE International Conference on Robotics and Automation (ICRA)}.\hskip 1em plus 0.5em minus 0.4em\relax IEEE, 2024, pp. 14\,576--14\,582.

\bibitem{damani2021primal}
M.~Damani, Z.~Luo, E.~Wenzel, and G.~Sartoretti, ``{PRIMAL}$ _2 $: Pathfinding via reinforcement and imitation multi-agent learning-lifelong,'' \emph{IEEE Robotics and Automation Letters}, vol.~6, no.~2, pp. 2666--2673, 2021.

\bibitem{wang2023scrimp}
Y.~Wang, B.~Xiang, S.~Huang, and G.~Sartoretti, ``{SCRIMP}: Scalable communication for reinforcement-and imitation-learning-based multi-agent pathfinding,'' in \emph{2023 IEEE/RSJ International Conference on Intelligent Robots and Systems (IROS)}.\hskip 1em plus 0.5em minus 0.4em\relax IEEE, 2023, pp. 9301--9308.

\bibitem{li2022multi}
W.~Li, H.~Chen, B.~Jin, W.~Tan, H.~Zha, and X.~Wang, ``Multi-agent path finding with prioritized communication learning,'' in \emph{2022 International Conference on Robotics and Automation (ICRA)}.\hskip 1em plus 0.5em minus 0.4em\relax IEEE, 2022, pp. 10\,695--10\,701.

\bibitem{li2020graph}
Q.~Li, F.~Gama, A.~Ribeiro, and A.~Prorok, ``Graph neural networks for decentralized multi-robot path planning,'' in \emph{2020 IEEE/RSJ International Conference on Intelligent Robots and Systems (IROS)}.\hskip 1em plus 0.5em minus 0.4em\relax IEEE, 2020, pp. 11\,785--11\,792.

\bibitem{li2021message}
Q.~Li, W.~Lin, Z.~Liu, and A.~Prorok, ``Message-aware graph attention networks for large-scale multi-robot path planning,'' \emph{IEEE Robotics and Automation Letters}, vol.~6, no.~3, pp. 5533--5540, 2021.

\bibitem{nguyen2020deep}
T.~T. Nguyen, N.~D. Nguyen, and S.~Nahavandi, ``Deep reinforcement learning for multiagent systems: A review of challenges, solutions, and applications,'' \emph{IEEE transactions on cybernetics}, vol.~50, no.~9, pp. 3826--3839, 2020.

\bibitem{kiran2021deep}
B.~R. Kiran, I.~Sobh, V.~Talpaert, P.~Mannion, A.~A. Al~Sallab, S.~Yogamani, and P.~P{\'e}rez, ``Deep reinforcement learning for autonomous driving: A survey,'' \emph{IEEE Transactions on Intelligent Transportation Systems}, vol.~23, no.~6, pp. 4909--4926, 2021.

\bibitem{mikolov2013efficient}
T.~Mikolov, K.~Chen, G.~Corrado, and J.~Dean, ``Efficient estimation of word representations in vector space,'' \emph{arXiv preprint arXiv:1301.3781}, 2013.

\bibitem{shannon1948mathematical}
C.~E. Shannon, ``A mathematical theory of communication,'' \emph{The Bell system technical journal}, vol.~27, no.~3, pp. 379--423, 1948.

\bibitem{kipf2016semi}
T.~N. Kipf and M.~Welling, ``Semi-supervised classification with graph convolutional networks,'' \emph{arXiv preprint arXiv:1609.02907}, 2016.

\bibitem{zhang2018link}
M.~Zhang and Y.~Chen, ``Link prediction based on graph neural networks,'' \emph{Advances in Neural Information Processing Systems}, vol.~31, 2018.

\bibitem{Bar-Gera2021}
\BIBentryALTinterwordspacing
H.~Bar-Gera, ``Transportation network test problems,'' Last Accessed July 19, 2024. [Online]. Available: \url{https://github.com/bstabler/TransportationNetworks}
\BIBentrySTDinterwordspacing

\bibitem{guo2022navigation}
H.~Guo, Z.~He, C.~Gao, and D.~Rus, ``Navigation with time limits in transportation networks: A fourth moment approach,'' \emph{IEEE Transactions on Intelligent Transportation Systems}, vol.~23, no.~12, pp. 23\,781--23\,796, 2022.

\bibitem{prakash2020algorithms}
A.~A. Prakash, ``Algorithms for most reliable routes on stochastic and time-dependent networks,'' \emph{Transportation Research Part B: Methodological}, vol. 138, pp. 202--220, 2020.

\bibitem{chen2017most}
B.~Y. Chen, C.~Shi, J.~Zhang, W.~H. Lam, Q.~Li, and S.~Xiang, ``Most reliable path-finding algorithm for maximizing on-time arrival probability,'' \emph{Transportmetrica B: Transport Dynamics}, vol.~5, no.~3, pp. 248--264, 2017.

\bibitem{guo2022dual}
H.~Guo, R.~Shi, D.~Rus, and W.-Y. Yau, ``Dual dynamic programming for the mean standard deviation {Canadian} traveller problem,'' \emph{IEEE Transactions on Vehicular Technology}, vol.~71, no.~11, pp. 11\,465--11\,479, 2022.

\bibitem{gao2021gp4}
C.~Gao, H.~Guo, and W.~Sheng, ``{GP4}: Gaussian process proactive path planning for the stochastic on time arrival problem,'' \emph{IEEE Transactions on Vehicular Technology}, vol.~70, no.~10, pp. 9849--9862, 2021.

\end{thebibliography}

% \section{Biography Section}
% If you have an EPS/PDF photo (graphicx package needed), extra braces are
%  needed around the contents of the optional argument to biography to prevent
%  the LaTeX parser from getting confused when it sees the complicated
%  $\backslash${\tt{includegraphics}} command within an optional argument. (You can create
%  your own custom macro containing the $\backslash${\tt{includegraphics}} command to make things
%  simpler here.)
 
% \vspace{11pt}

% \bf{If you include a photo:}\vspace{-33pt}
% \begin{IEEEbiography}[{\includegraphics[width=1in,height=1.25in,clip,keepaspectratio]{fig1}}]{Michael Shell}
% Use $\backslash${\tt{begin\{IEEEbiography\}}} and then for the 1st argument use $\backslash${\tt{includegraphics}} to declare and link the author photo.
% Use the author name as the 3rd argument followed by the biography text.
% \end{IEEEbiography}

% \vspace{11pt}

% \bf{If you will not include a photo:}\vspace{-33pt}
% \begin{IEEEbiographynophoto}{John Doe}
% Use $\backslash${\tt{begin\{IEEEbiographynophoto\}}} and the author name as the argument followed by the biography text.
% \end{IEEEbiographynophoto}

\vfill

\end{document}